\documentclass[10pt,conference]{IEEEtran}

\usepackage[utf8]{inputenc}
\usepackage[T1]{fontenc}
\usepackage{placeins}
\usepackage{graphicx}
\usepackage{url}
\usepackage{subcaption}
\usepackage{caption}
\usepackage{booktabs}   
\usepackage{array}      
\usepackage{amsmath}
\usepackage{amssymb}
\usepackage{float}
\usepackage{acronym}   
\usepackage{siunitx}   
\acrodef{CA}{cellular automaton}
\acrodefplural{CA}[CAs]{cellular automata}
\acrodef{NCA}{Neural Cellular Automaton}
\acrodefplural{NCA}[NCAs]{Neural Cellular Automata}
\acrodef{NCPU}{Neural Cellular Processing Unit}
\acrodef{ALU}{arithmetic logic unit}
\acrodef{LSB}{least-significant bit}
\acrodef{MSB}{most-significant bit}
\acrodef{MSE}{mean-squared error}
\acrodef{CNN}{convolutional neural network}
\acrodefplural{CNN}[CNNs]{convolutional neural networks}
\acrodef{MLP}{multilayer perceptron}
\acrodefplural{MLP}[MLPs]{multilayer perceptrons}

\newcommand{\mycaptionsize}{scriptsize}

\usepackage{eso-pic}
\AddToShipoutPictureBG*{\AtPageLowerLeft{\put(\LenToUnit{0.5\paperwidth},\LenToUnit{0.35in}){\makebox[0pt]{\parbox{0.86\paperwidth}{\centering\scriptsize
Accepted at the 2027 IEEE Symposium Series on Computational Intelligence (SSCI 2027). \copyright~2027 IEEE. Personal use of this material is permitted. Permission from IEEE must be obtained for all other uses, in any current or future media, including reprinting/republishing this material for advertising or promotional purposes, creating new collective works, for resale or redistribution to servers or lists, or reuse of any copyrighted component of this work in other works.}}}}}
\begin{document}
\title{NCPU: Towards Cell-Inspired Spatial Computing}
\author{\IEEEauthorblockN{Iliya Zhechev}
\IEEEauthorblockA{Sofia University, Bulgaria\\
\texttt{iliya.zhechev@gmail.com}}
\and
\IEEEauthorblockN{Piotr Walas}
\IEEEauthorblockA{Warsaw University of Technology, Poland\\
\texttt{Piotr.Walas@pw.edu.pl}}}
\maketitle
\begin{abstract}
Simple \acp{CA} can be computationally universal, but their rules are designed by hand. We train \acp{NCA}, which learn a local update rule by gradient descent, to compute exact digital logic. Bits are encoded as clusters of cells on a 2D grid, and a result reaches the output only by propagating through intermediate cells. One model is trained per task: logic gates, 4- and 8-bit ripple-carry adders, and a preliminary 8-bit \ac{ALU}. All gate and adder models reach \qty{100}{\percent} bit correctness in 10 of 10 seeds, and exhaustive tests over all input pairs find at most 1 error in 256 (4-bit) and 7 in \num{65536} (8-bit). Adders trained on operands of up to 3 bits generalize to 8 bits. The adders propagate the carry as a wave, with a delay of about 3.5 steps per bit of the carry chain. Outputs are correct only within the trained rollout length.
\end{abstract}
\acresetall 

\section{Introduction}
\label{sec:intro}
Can a rule learned by gradient descent, applied identically at every cell of a grid and seeing only its neighbours, compute exact digital logic? A \ac{CA} is a discrete dynamical system in which one local rule updates every cell of a grid. Simple \acp{CA}, such as Conway's Game of Life~\cite{ref_conways_life}, are computationally universal, but their rules, and the initial states that make them compute, are designed by hand. \acp{NCA}~\cite{ref_Growing_Neural_Cellular_Automata} replace the fixed rule with a neural network trained from initial and target grid states; the trained model updates the grid as a \ac{CA} does.

Most \ac{NCA} work targets pattern growth and classification, where each region of the output depends mostly on nearby cells. In digital logic, an output bit can depend on input bits that are far away on the grid. Addition is a simple case of this: a carry can pass through every bit of the operands. The \acp{NCA} we train solve it in a way that resembles a ripple-carry adder: the carry moves as a wave from bit to bit. Each step of the wave is the same local operation, so the models also add operands wider than those seen in training.

We study \acp{NCA} as a \ac{NCPU}, a substrate trained to perform digital logic spatially. The goal is not to move all computation to an \ac{NCPU}, but to find out whether spatial \ac{NCA} computation is possible and what it requires. We train a separate \ac{NCA} model per task over a sequence of tasks of increasing complexity, from logic gates to ripple-carry adders and a preliminary \ac{ALU}, encoding all inputs and outputs as spatial bit patterns on a 2D grid, and an ablation study examines how model hyperparameters affect the dynamics.

This question matters in two ways. For artificial life, exact logic is a strict test of what self-organizing local rules can do: an error flips a bit, so success and failure are unambiguous, and the mechanisms the rule finds, such as the carry wave of Section~\ref{sec:carry}, show how a global function can come from local interactions. For unconventional computing, an \ac{NCA} that implements arithmetic primitives is a step towards trainable hardware built from identical local units.

Our contributions are: (1) \ac{NCA}-based digital computation across a sequence of tasks from single gates to 8-bit adders, one model per task, and a preliminary \ac{ALU}; (2) length generalization from smaller to larger adders; and (3) a carry-wave mechanism that emerges from spatial propagation constraints, measured by its propagation delay and by channel ablation. The code and data for this study are available at \url{https://github.com/ichko/ncpu}, and an interactive demo at \url{https://ncpu.pages.dev}.

\section{Related Work}

Learned update rules extend classical \acp{CA} in several directions. \ac{NCA} was introduced by a Google research group, which showed that learned rules sustain and regenerate damaged patterns~\cite{ref_Growing_Neural_Cellular_Automata}; later work showed \ac{NCA} can generalize matrix computations, suggesting a building block for universal computation~\cite{ref_universal_nca}. Others control \ac{NCA} dynamics through mass conservation~\cite{ref_MaCE}, asynchrony~\cite{ref_nca_async} and isotropy~\cite{ref_nca_isotropic}, and differentiable logic cellular automata bridge rule-based automata and learned pattern generation~\cite{ref_difflogic_ca}. \acp{NCA} have been applied to MNIST classification~\cite{ref_nca_mnist}, signal response~\cite{ref_nca_signals}, distributed sensing~\cite{ref_nca_sensing_hardware} and physical 2D shape classification~\cite{ref_physical_nca_walker}.

\acp{NCA} have been implemented on embedded chips~\cite{ref_nca_hardware_embedded}, and memristor~\cite{ref_memristor_ca} and photonic~\cite{ref_photonic_eca_2023} substrates have been explored, suggesting that an \ac{NCPU}-like unit could in principle be realized in low-power hardware.

Relative to Béna et al.~\cite{ref_universal_nca}, who study whether \ac{NCA} can approximate \emph{universal} computation, our contribution is complementary: we do not pursue universality, but ask whether \ac{NCA} dynamics can realize exact ripple-carry arithmetic, and what emergent mechanisms this induces.

Closest to our work, Wei~\cite{ref_sead} derives an \ac{NCA} architecture (SEAD) from locality, symmetry and stability constraints, iterates local convolutional rules until they converge to an attractor, and reports length generalization on parity and multi-digit addition, and the learning of Rule~110. That work concentrates on length generalization of single operations. We instead use the standard \ac{NCA} formulation~\cite{ref_Growing_Neural_Cellular_Automata} and ask which digital building blocks it can realize: logic gates, ripple-carry adders with an explicit carry-out, and an opcode-controlled \ac{ALU}, with bits encoded as spatial clusters and inputs and outputs in separate regions of a 2D grid. We also analyze the mechanism that a trained adder uses (the carry wave) and where the approach fails (\ac{ALU} carry-out).

\section{Methods}
\label{sec:methods}

\subsection{Setup}

The \ac{NCPU} model pipeline begins with a \textit{perception stage}, in which Sobel filtering is applied to the current grid state. The resulting perception states are subsequently passed to the \textit{N\ac{CA} update stage}, where the learned \ac{NCA} transition rule is applied. This is followed by an \textit{alive masking} step that deactivates cells whose state values fall below a predefined threshold. Finally, to promote generalization, state updates are stochastically masked during a \textit{stochastic update stage}, in which each cell's update is applied with a fixed probability, the \emph{fire rate}.

All four stages follow the standard \ac{NCA} formulation~\cite{ref_Growing_Neural_Cellular_Automata}, with one deviation: we disable alive masking. Section~\ref{sec:ablation} shows that masking prevents convergence on these tasks, so every experiment reported below runs with it off (alive masking set to 0.0).

Additionally, all experiments use 16 channels and the standard learned transition network~\cite{ref_Growing_Neural_Cellular_Automata}, a single hidden layer of 128 units applied to three Sobel-based perception channels, with fire rate $0.5$ and a \numproduct{7 x 7} kernel for the adder and \ac{ALU} experiments, or \numproduct{5 x 5} for the gate experiments (Section~\ref{sec:ablation} reports the kernel sweep that motivated the larger kernel).

Optimization uses Adam with learning rate \num{e-4}, batch size 8 (16 for the gate experiments), and a rollout length sampled uniformly per optimization step: \numrange{32}{64} steps for gates, \numrange{64}{128} for adders and length generalization, and \numrange{48}{96} for the \ac{ALU}.

The grid is fixed within each experiment: input and output bit clusters (Section~\ref{sec:experiment_setup}) are placed on a fixed spatial layout that packs more clusters into the grid as bit-width grows, rather than resizing the grid, and the layout is bottom-aligned so that the \ac{LSB} occupies the same absolute position at every bit-width. Grid size is selected per task to fit the largest encoding it must hold within practical memory constraints: \numproduct{28 x 48} for single gates, \numproduct{80 x 112} for the 4-bit adder (both layouts), the two-column 8-bit adder and all length-generalization runs, \numproduct{80 x 192} for the single-column 8-bit adder, and \numproduct{96 x 112} for the \ac{ALU}.

\subsection{Experimental setup}
\label{sec:experiment_setup}

The \ac{NCPU} operates on a two-dimensional grid with cell values in a bipolar range $[-1, 1]$, so that logical low and high sit at the opposite extremes ($-1$ and $+1$ respectively), while dead cells remain at $0$, avoiding the bias a unipolar $[0,1]$ range would introduce.

A bit is carried not by a single cell but by a \emph{bit cluster}: a disc of radius $4$ cells in which every cell holds the same value, $+1$ or $-1$. The bit is read as the sign of the mean activation over the disc, so noise in individual cells does not change the value, and the cluster gives the local rule an area through which signals can be routed.

To enforce computation propagation, the grid is divided into input (left) and output (right) regions, so the model must propagate and transform information from left to right. Initial states contain only input bits, while target states define only output bits. The number and arrangement of bits vary by task, from logic gates to an \ac{ALU}.

\subsection{Loss Functions}
\label{sec:loss}

For \ac{NCPU}, we use a \textbf{masked \ac{MSE} loss}, computed over selected grid cells and rollout frames, allowing targeted supervision of output regions during training.

\begin{equation}
\mathcal{L}_{\text{masked}} = \frac{1}{B\,\lvert\mathcal{T}\rvert\,\sum_i m_i}
\sum_{b=1}^{B}\sum_{t \in \mathcal{T}}\sum_{i} m_i\left(\hat{y}_{b,t,i} - y_{b,i}\right)^2
\end{equation}

The target $y$ is a grid of the same size as the state. Inside the disc of each output bit, every cell is set to $+1$ (\num{0.99} in the implementation, from 8-bit pixel values) if the bit is 1 and to $-1$ if it is 0. All cells outside the output discs are set to $0$ and have $m_i = 0$, so they receive no supervision. The input bits are written in the same way, inside their input discs, into channels~0 and~1 of the initial state; all other cells and channels start at $0$. Channel~1 is read-only, so the input stays visible during the whole rollout, while channel~0, which also carries the output, is free to change.

\noindent In the loss, $b$ indexes the $B$ samples in a batch, $t$ indexes the supervised rollout frames $\mathcal{T}$ (every frame of the rollout for the gate and adder experiments, the last eight for the \ac{ALU}), $i$ indexes the $N$ grid cells, $\hat{y}_{b,t,i}$ is the channel-0 activation of cell $i$ at frame $t$, $y_{b,i}$ is the target activation of cell $i$, and $m_i \in \{0,1\}$ is $1$ if and only if cell $i$ lies in the output region, that is, inside one of the output-bit clusters. Since $\lvert\mathcal{T}\rvert$ differs by task family, loss values and convergence steps are comparable within a family but not directly across families.

All internal dynamics, including the carry wave of Section~\ref{sec:carry}, receive no direct supervision.

\subsection{Metrics}
The \ac{NCPU} models are evaluated against the following metrics. \textbf{Bit correctness} is the fraction of output bits matching the target, averaged over the batch. A bit is counted as correct when the sign of the mean channel-0 activation of its output cluster matches the target bit (positive for logical 1, negative for logical 0) on the final rollout frame $T$. Test inputs are drawn uniformly at random from the valid inputs for each task. For the \ac{ALU} this is 64 samples per operation, so a single-bit accuracy carries a \qty{95}{\percent} interval of about $\pm\num{12.5}$ percentage points. Each evaluation runs a fresh rollout of $T$ steps from the test input; no test input is reused.

\textbf{Convergence speed} is the number of training steps required to reach a given bit-correctness threshold (\qty{100}{\percent} unless stated otherwise), measured over 10 seeds per task; we report mean~$\pm$~standard deviation.

\section{Ablation}
\label{sec:ablation}
\subsection{Convergence-space sweep}
To select hyperparameters, we performed a grid search across convolution kernel size, alive masking threshold, and weight initialization on the single-gate task, measuring normalized convergence speed. During ablation we considered the model to have converged when it reached \qty{100}{\percent} bit correctness. The convergence speed is presented on a reversed scale from 1.0 to 0.0, where 1.0 means convergence within the first \num{2000} steps and 0.0 means convergence was not achieved.

The model is particularly sensitive to alive masking: disabling it yields the best convergence, while enabling it produces almost zero successful runs. This is a \emph{negative} result: it contradicts the original Growing \ac{NCA} formulation~\cite{ref_Growing_Neural_Cellular_Automata}, where masking stabilizes pattern growth, because here masking deactivates the intermediate cells through which left-to-right propagation must travel. We therefore disabled it in all subsequent experiments. The sweep used the single-gate task; we expect the effect to grow with circuit depth, since masking removes the near-zero intermediate cells that a longer path, such as the 8-bit carry chain, must cross.

Kernel size has the second largest influence: larger kernels consistently reduce convergence time, suggesting that a wider receptive field gives each cell more context. A kernel size of 7 is a practical trade-off between convergence speed and computational cost, and we use it for the adder and \ac{ALU} experiments; the gate runs of Section~\ref{sec:results} use the \numproduct{5 x 5} trainer default.

The model also prefers random over zero weight initialization, and a higher fire rate tends to help, which may matter for more complex circuits.

\begin{figure}[h!]
\centering

\captionsetup{justification=centering, font=\mycaptionsize}
\centering
\includegraphics[width=\linewidth]{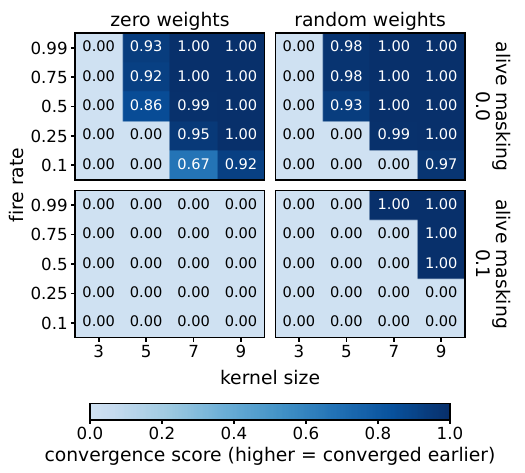}    
\caption{Convergence on the single-gate task for alive-masking thresholds 0.0 and 0.1 (rows; at 0.3 no run converged), weight initialization (columns), fire rate and kernel size. Darker blue means faster convergence: 1.0 is convergence within the first \num{2000} steps, 0.0 is no convergence.}
\label{fig:heatmaps}
\end{figure}

\section{Results}
\label{sec:results}

We evaluate \ac{NCPU} on a sequence of tasks of increasing complexity, training one model per task: single logic gates, ripple-carry adders, length generalization, and a preliminary 8-bit \ac{ALU}. All experiments use the masked \ac{MSE} loss (Section~\ref{sec:loss}).
\subsection{Logic gates}
We trained \ac{NCPU} on six standard two-input gates (AND, OR, NOR, NAND, XOR, XNOR) and two slightly more complex circuits (half-adder and majority gate). Each experiment was trained over 10 random seeds; convergence speed and bit accuracy are reported as mean~$\pm$~standard deviation across seeds.

\begin{table}[htbp]
\centering
\caption{Multi-seed convergence results for logic gate, adder, and \ac{ALU} experiments. Convergence steps are optimization steps to first reach \qty{100}{\percent} bit correctness. Every gate and adder value is the mean over 10 seeds per task, and ``$\pm$'' is the population standard deviation (ddof$=0$) across those seeds; the \ac{ALU} is a single non-converged run and is not seed-averaged. Final bits are correct output bits on the last rollout frame out of a per-task maximum of 1 (gates, majority3), 2 (half-adder), 5 (4-bit) or 9 (8-bit). \emph{cols1} and \emph{cols2} denote the single- and two-column operand layouts of Section~\ref{sec:adders}; the 4-bit cols2 and 8-bit cols1 rows are the layout control, trained for \num{20000} steps.}
\label{tab:gate_results}
\begin{tabular}{lccc}
\toprule
Task & Seeds & Convergence steps & Final bits \\
\midrule
AND       & 10 & \num{49 \pm 35} & $1.00 \pm 0.00$ \\
NAND      & 10 & \num{45 \pm 21} & $1.00 \pm 0.00$ \\
NOR       & 10 & \num{35 \pm 12} & $1.00 \pm 0.00$ \\
OR        & 10 & \num{54 \pm 36} & $1.00 \pm 0.00$ \\
XNOR      & 10 & \num{67 \pm 23} & $1.00 \pm 0.00$ \\
XOR       & 10 & \num{85 \pm 42} & $1.00 \pm 0.00$ \\
Half-adder & 10 & \num{366 \pm 138} & \num{2.00 \pm 0.10} \\
Majority3 & 10 & \num{116 \pm 53} & $1.00 \pm 0.00$ \\
4-bit adder (cols1) & 10 & \num{9976 \pm 1342} & $5.00 \pm 0.00$ \\
4-bit adder (cols2) & 10 & \num{3770 \pm 824} & $5.00 \pm 0.00$ \\
8-bit adder (cols2) & 10 & \num{5142 \pm 1035} & $9.00 \pm 0.00$ \\
8-bit adder (cols1) & 10 & \multicolumn{2}{c}{Not converged (\num{20000} steps)} \\
\ac{ALU} (8-bit) & 1 & \multicolumn{2}{c}{Not converged (\num{600000} steps)} \\
\bottomrule
\end{tabular}
\end{table}

\begin{figure}[t]
    \centering
    \captionsetup{font={\mycaptionsize}}
    \includegraphics[width=\linewidth]{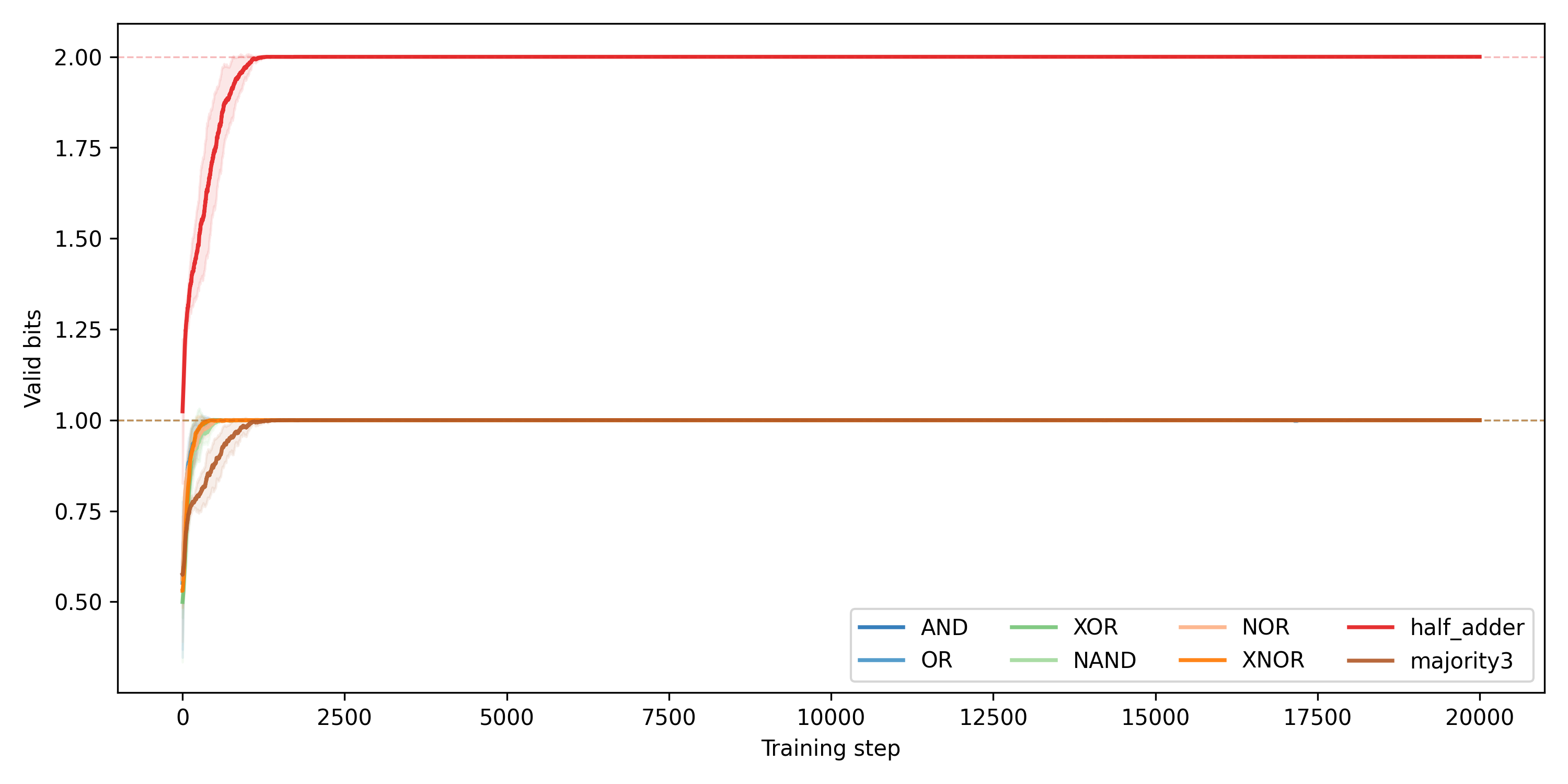}
\caption{Training accuracy (correct output bits) across seeds for all logic gates, half-adder, and majority3. Linearly separable gates converge faster than non-linear ones.}
\label{fig:gate_seeds}
\end{figure}

Linearly separable gates (AND, NAND, NOR, OR) converge in under 55 steps, non-linearly separable ones (XOR, XNOR) in \numrange{67}{85}, and the three-input majority gate in about 116; the half-adder, which must produce both sum and carry, needs about 366. The same ordering is visible across seeds (Fig.~\ref{fig:gate_seeds}). The adders also converge fully (5/5 and 9/9 bits, Table~\ref{tab:gate_results}) but need far more optimization steps; the \ac{ALU} has not converged after \num{600000} steps.

\begin{figure}[h!]
    \centering
    \captionsetup{font={\mycaptionsize}}
    \begin{subfigure}{0.47\linewidth}
        \centering
        \includegraphics[width=\linewidth]{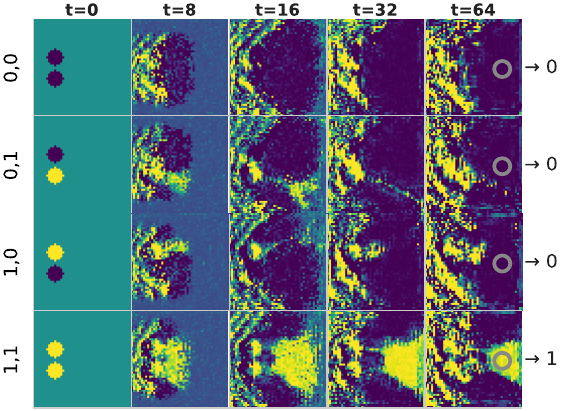}
    \end{subfigure}
    \hfill
    \begin{subfigure}{0.47\linewidth}
        \centering
        \includegraphics[width=\linewidth]{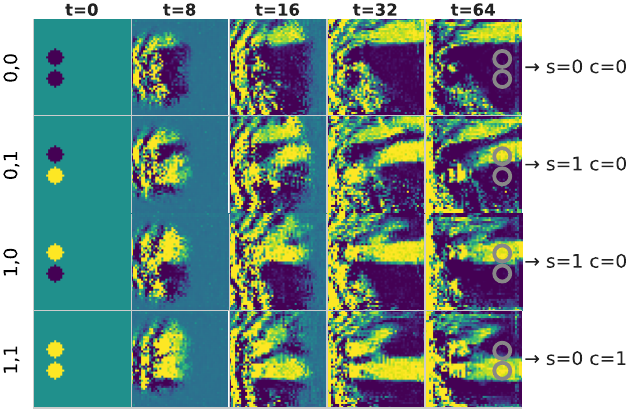}
    \end{subfigure}
\caption{Channel~0 rollouts for the AND gate (left) and the half-adder (right). Circles indicate output bit locations. The chaotic region on the left may indicate wave-based computation (not tested).}
\label{fig:single_gate_rollouts}
\end{figure}

Example rollouts for the AND gate and the half-adder (Fig.~\ref{fig:single_gate_rollouts}) show clear propagation routes. They also show a highly chaotic region on the left of each grid. We interpret this as a sign of wave-based computation, but we have not tested this interpretation and report it as an observation only.

\subsection{Ripple-carry adders}
\label{sec:adders}

\begin{figure}[h!]
    \centering
    \captionsetup{font={\mycaptionsize}}
    \begin{subfigure}[b]{0.47\linewidth}
        \centering
        \includegraphics[width=\linewidth]{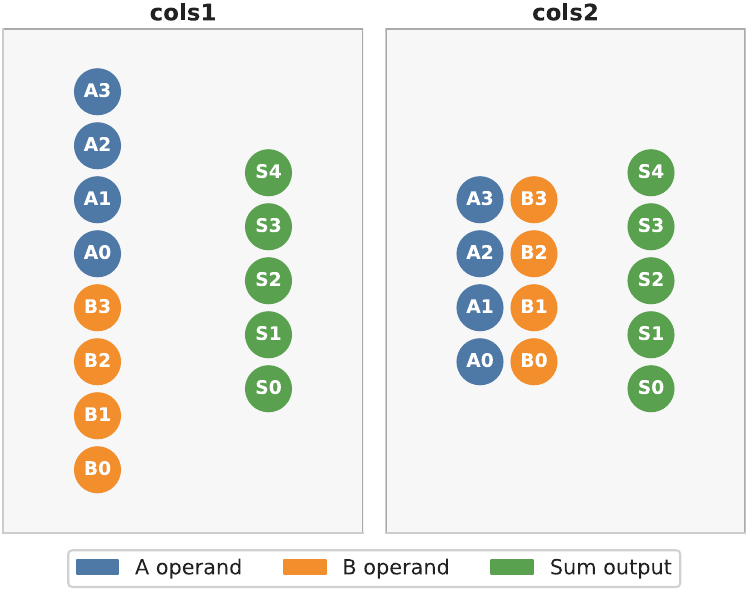}
        \caption{Input layouts}
        \label{fig:4bit_layouts}
    \end{subfigure}
    \hfill
    \begin{subfigure}[b]{0.40\linewidth}
        \centering
        \includegraphics[width=\linewidth]{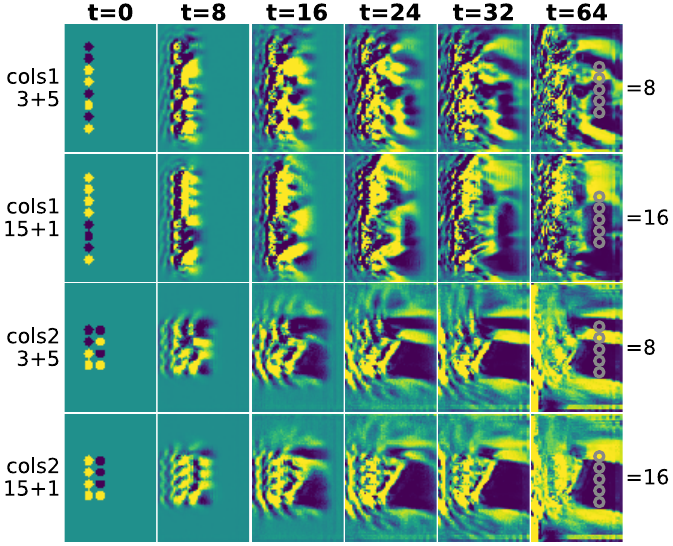}
        \caption{Channel~0 rollouts}
        \label{fig:4bitAdd}
    \end{subfigure}
    \caption{4-bit adder: single-column (\emph{cols1}) versus two-column (\emph{cols2}) input layouts (a), and Channel~0 rollouts on the same two input combinations for both layouts (b). Top two rows show cols1; bottom two rows show cols2.}
    \label{fig:adder_results}
\end{figure}

We trained ripple-carry addition at two scales (4-bit and 8-bit, 10 seeds each), producing a sum and a carry-out. The 4-bit case compared a single-column layout (\emph{cols1}) with a two-column layout (\emph{cols2}), with operands A and B in adjacent columns (Fig.~\ref{fig:4bit_layouts}). cols2 converges in \num{3770 \pm 824} steps, 2.6 times faster than cols1 (\num{9976 \pm 1342}); Channel~0 rollouts (Fig.~\ref{fig:4bitAdd}) show cols2 settling its output column earlier. The two-column 8-bit adder converged in \num{5142 \pm 1035} steps. To separate layout from operand width, we also trained the 8-bit adder in the single-column layout: none of 10 seeds converged within \num{20000} steps (best 6.4 to 7.0 of 9 bits; one seed run to \num{50000} steps reached 6.75). At a fixed width, the layout therefore decides convergence.

\subsection{Exactness}
\label{sec:exact}

Bit correctness during training is measured on random inputs. To test exactness, we evaluated trained adders on every input pair at step 128, using models re-trained with the same code: ten 4-bit cols2 models, three 4-bit cols1 models and two 8-bit cols2 models. Eleven of the 13 4-bit models are correct on all 256 pairs and the other two fail on one pair each; the two 8-bit models fail on 7 and 6 of \num{65536} pairs (\qty{99.99}{\percent} correct). The results are the same when every step from 64 to 128, the rollout range used in training, must be correct, except for three more 8-bit pairs. Beyond that range the answer is not stable: run to \num{1024} steps, three of four models tested lose most answers by step 256 (\qtyrange{2}{69}{\percent} still correct), and the fourth keeps half of them to step \num{1024}. The models compute the sum within the trained horizon but do not hold it as a fixed point.

\subsection{Length generalization}

\begin{figure}[h!]
    \centering
    \captionsetup{font={\mycaptionsize}}
    \begin{subfigure}{0.30\linewidth}
        \centering
        \includegraphics[width=\linewidth]{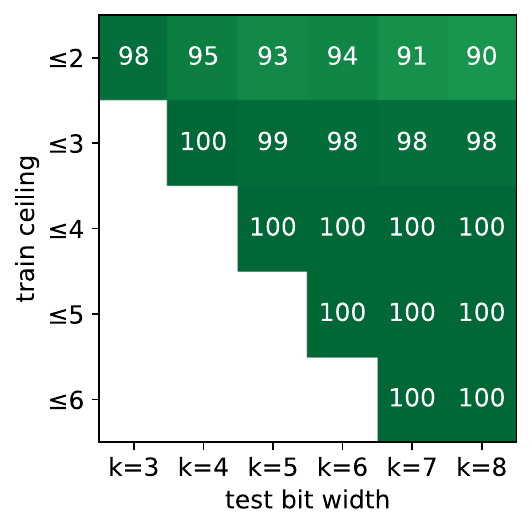}
    \end{subfigure}
    \hfill
    \begin{subfigure}{0.66\linewidth}
        \centering
        \includegraphics[width=\linewidth]{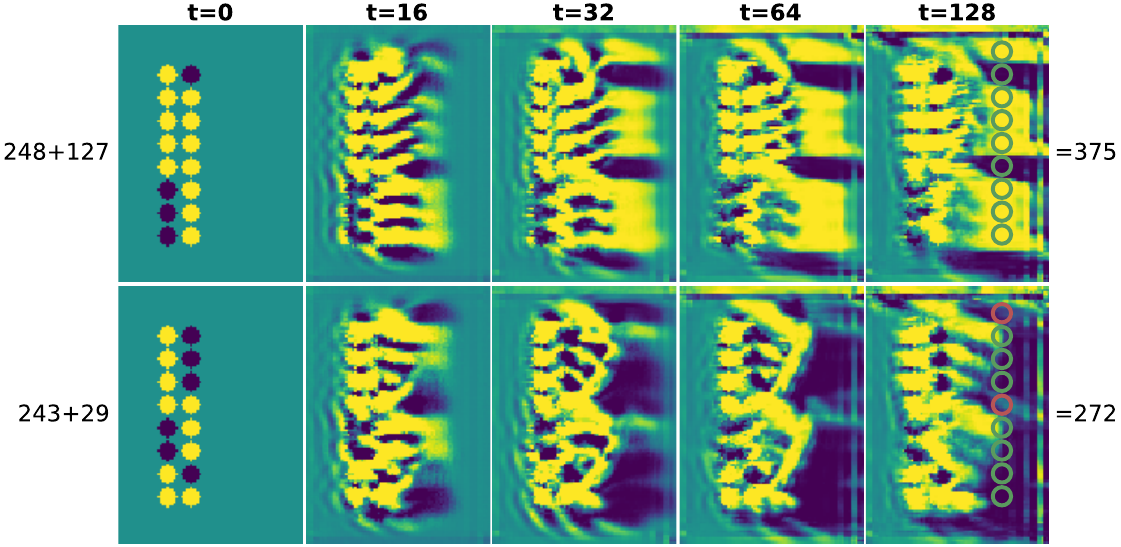}
    \end{subfigure}
    \caption{Length generalization results. \textit{Left}: accuracy (\unit{\percent}) for each training ceiling $K$ and test bit width. \textit{Right}: success (top) and failure (bottom) for a $K{=}3$ model on 8-bit input.}
    \label{fig:extrap_results}
\end{figure}

To test whether the model can be trained on smaller adders and generalize to larger ones without encountering them during training, we trained models on $K$-bit adders ($K \in \{2, 3, 4, 5, 6\}$) on the fixed \numproduct{80 x 112} grid, then evaluated them at test bit widths $K+N$ for $N \geq 1$. Bit clusters are placed from the bottom of the grid upward, at the same positions for every bit width. A model trained on $K$ bits uses only the lower part of the grid. At test time we add bits above the trained ones and check whether the learned rule extends to them. Models trained on $K \geq 3$ bits extrapolate near-perfectly to all widths up to 8, achieving at least \qty{98}{\percent} bit correctness regardless of the gap. Training on only $K = 2$ bits still works but degrades proportionally to the gap. A sharp transition is visible at $K = 3$: below it generalization degrades with the gap; above it accuracy is insensitive to the gap (Fig.~\ref{fig:extrap_results}).

We hypothesize, without a direct test, that $K=3$ matters because it is the smallest width with an intermediate bit between the \ac{LSB} and the \ac{MSB}. With $K=2$ the model never propagates a carry through an intermediate position, so the positions added at test time are new to it; with $K \geq 3$ it has learned that step, and packing more positions into the grid re-applies the same local rule.

\subsection{Arithmetic and logic unit}
\label{sec:alu}

\begin{figure}[t]
    \centering
    \captionsetup{font={\mycaptionsize}}
    \begin{subfigure}[b]{0.30\linewidth}
        \centering
        \includegraphics[width=\linewidth]{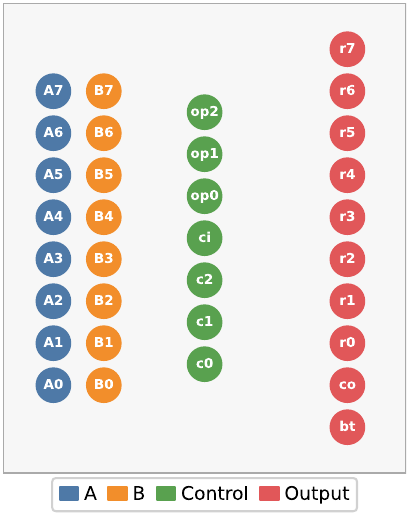}
        \caption{Spatial layout of the 8-bit \ac{ALU} grid.}
        \label{fig:alu_layout}
    \end{subfigure}
    \hfill
    \begin{subfigure}[b]{0.64\linewidth}
        \centering
        \includegraphics[width=\linewidth]{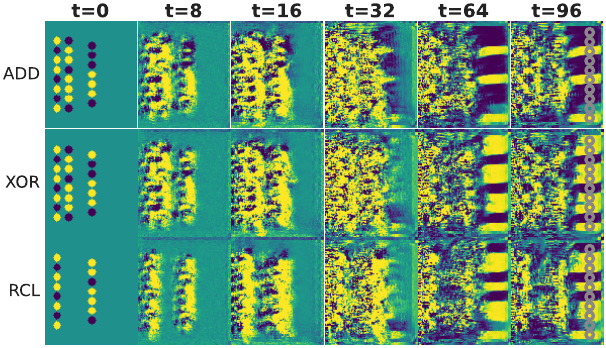}
        \caption{Channel~0 rollout for ADD (carry-heavy), XOR (bitwise), and RCL (rotate-through-carry). Gray circles mark output-bit locations.}
        \label{fig:alu_rollout}
    \end{subfigure}
\caption{8-bit \ac{ALU} experiment: spatial layout (a) and Channel~0 rollouts for three representative operations (b).}
\label{fig:ALU}
\end{figure}

The \ac{ALU} experiment is preliminary: it is a single run of \num{600000} optimization steps, which we could not repeat with other seeds within our compute budget. The model did not reach \qty{100}{\percent} bit correctness. We report this run because it shows where the approach succeeds and where it breaks down, but every \ac{ALU} finding below holds for this one run only and is not a converged or seed-averaged result. The layout uses two input columns for the operands $A$ and $B$, a control column carrying a 3-bit opcode together with carry-in and a 3-bit condition code, and a single output column on the right holding ten bits: the 8-bit result, carry-out, and branch-taken (Figs.~\ref{fig:alu_layout} and \ref{fig:alu_rollout}).

The \ac{ALU} supports eight operations: ADD ($A + B + c_{\text{in}}$), SUB ($A - B - c_{\text{in}}$; carry-out $= \text{NOT borrow}$), AND, OR, XOR, NOT (on $B$), RCL (rotate left through carry) and RCR (rotate right through carry). Carry-out is the arithmetic carry for ADD and SUB, the bit rotated out of the operand for RCL and RCR, and is defined as zero for the bitwise operations. The branch-taken flag is set when the result and the internally represented zero/negative/carry flags satisfy the 3-bit condition code (EQ, NE, CS, CC, MI, PL, AL, NV), so it requires the rule to evaluate a predicate over flags it must represent itself rather than to emit one stored bit.

\begin{figure}[t]
    \centering
    \captionsetup{font={\mycaptionsize}}
    \includegraphics[width=\linewidth]{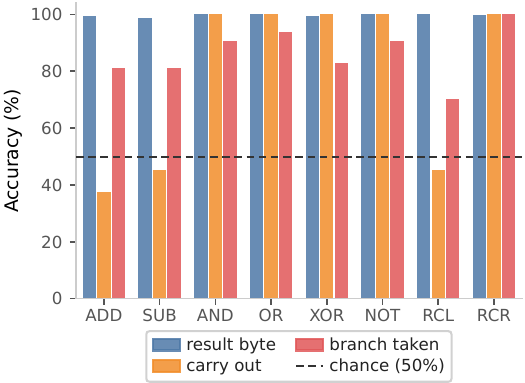}
    \caption{Per-operation accuracy of the 8-bit \ac{ALU} by output group: result bits, carry-out, branch-taken. Single non-converged run; see Section~\ref{sec:alu} for the chance-level caveats. The dashed line marks the \qty{50}{\percent} chance level of a single bit. The deficit is concentrated in carry-out on ADD, SUB and RCL.}
    \label{fig:ALU_OP}
\end{figure}

While result-bit accuracy is essentially perfect across all operations (at least \qty{98.6}{\percent} per operation, \qty{99.7}{\percent} mean), the auxiliary outputs reveal structured failure modes (Fig.~\ref{fig:ALU_OP}). Disaggregating by output isolates the deficit. Carry-out is a single bit, so chance is \qty{50}{\percent}; on ADD, SUB and RCL carry accuracy is \qtyrange{37}{45}{\percent}, and at 64 samples per operation the \qty{95}{\percent} interval is about $\pm\num{12.5}$ points, so carry is not learned on those operations, and we do not claim it is below chance. For AND, OR, XOR and NOT carry-out is defined to be zero, so their \qty{100}{\percent} is also what a rule that never emits a carry achieves; only RCR, whose carry-out is a genuine 50/50 bit (the \ac{LSB} of $B$), is informative, and there it reaches \qty{100}{\percent}. In this run, carry-out is therefore learned for one operation only, RCR, and not learned on three (ADD, SUB, RCL), while the remaining four are uninformative by construction. Branch-taken accuracy sits at \qty{86.3}{\percent} on average, above the \qty{50}{\percent} random baseline but well below the result-bit ceiling. RCL and RCR differ only in which operand bit is copied to carry-out (\ac{MSB} versus \ac{LSB}), yet fall on opposite sides of this split; with a single non-converged run we cannot tell whether that asymmetry is structural or an artifact of the run, and flag it as an open question.

\section{Discussion}
\label{sec:discussion}

\subsection{Baselines and scope}
\label{sec:baselines}

We scope this work as an existence proof. We ask whether a purely local, uniform \ac{NCA} update rule can be trained to perform structured digital logic under constraints a conventional network does not face: a \numproduct{7 x 7} neighborhood, no global addressing, and in-place computation in which information reaches the output only by propagating through intermediate cells. There is no directly comparable published baseline: most prior \ac{NCA} work targets pattern generation, and the closest work~\cite{ref_sead} studies length generalization of addition with a different \ac{NCA} architecture. A feedforward \ac{CNN} with the same \numproduct{7 x 7} receptive field would need one layer per propagation step, and so many times more parameters, to reach the outputs; our question is not whether some network can add, but whether one local rule shared by all cells can, and what dynamics that induces. The relevant controls are internal: the kernel-size and layout ablations, and the chance-level analysis above.

\subsection{Carry-wave}
\label{sec:carry}

\begin{figure}[h!]
    \centering
    \captionsetup{font={\mycaptionsize}}
    \includegraphics[width=\linewidth]{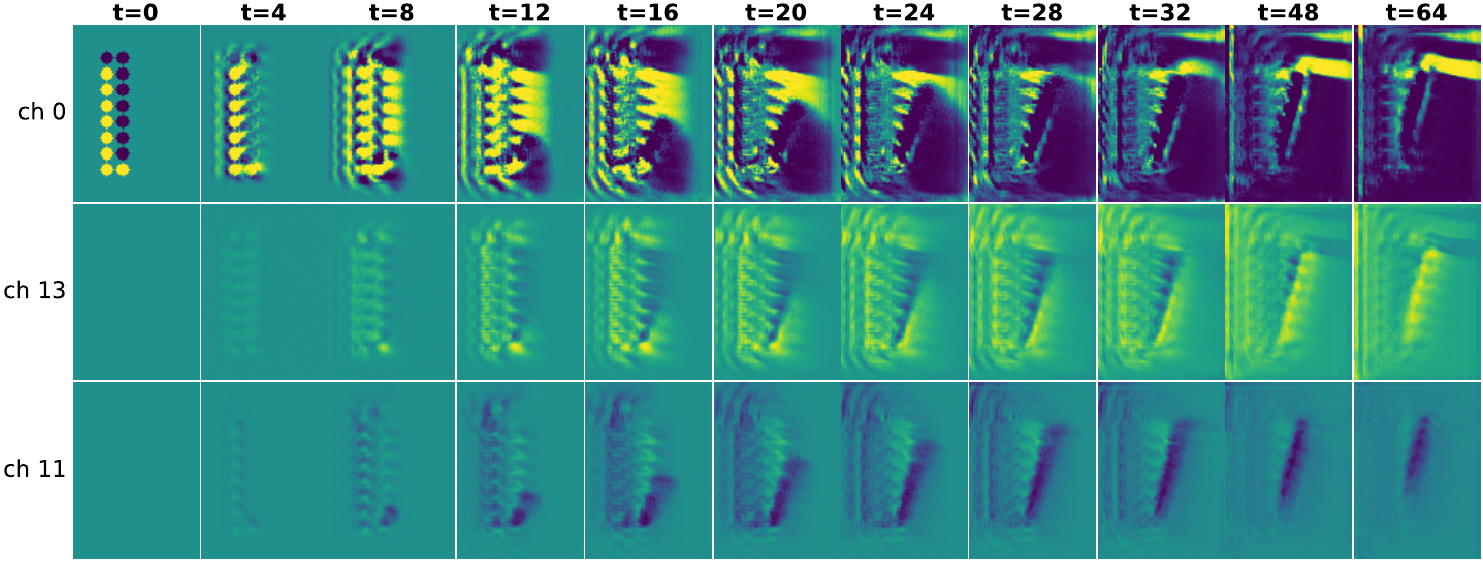}
    \caption{Hidden channels most strongly expressing the carry wave during the addition $127 + 1 = 128$ (7-bit carry chain). Each row is a single channel of the trained 8-bit adder; columns are \ac{NCA} timesteps. A wavefront propagates from the \ac{LSB} (bottom) toward the carry-out \ac{MSB} (top) over about 30 steps.}
    \label{fig:carry_wave}
\end{figure}

Although the \ac{NCA} update rule is local and applied uniformly, the 8-bit adder learns a \emph{sequential} strategy reminiscent of a classical ripple-carry adder. Running the trained model on $127 + 1 = 128$, which has a 7-bit carry chain, and inspecting the hidden channels reveals a wavefront that starts near the \ac{LSB} output and sweeps upward toward the carry-out \ac{MSB} over tens of timesteps (Fig.~\ref{fig:carry_wave}). Higher-order bits are effectively held neutral until the carry signal from below arrives, which is the same constraint a ripple-carry circuit imposes in silicon, yet no temporal structure is present in the training loss. The pattern is consistent with the spatial distance between output clusters imposing a minimum number of updates for information to travel.

We measured the wave on two 8-bit models and one 4-bit model. For each output bit we recorded the step after which it stays correct, and grouped bits by the length of the carry chain that ends at them (128 random inputs and worst-case chains). The settle step grows linearly with chain length, from 6--9 steps without a carry to 33--35 steps for a chain of 8 bits: each chain bit adds about 3.5 steps (2.6 and 3.5 for the two 8-bit models, 4.4 for the 4-bit model), the delay expected if the carry travels from one bit cluster to the next. To test whether the wave is causal, we set one hidden channel to zero at every step. Most channels are needed for every addition, but in each model one channel is specific to carries. Without channel~5 of the first 8-bit model, \qty{90}{\percent} of the additions without a carry stay correct, but only \qty{18}{\percent} of those with a chain of three or more bits; for channel~11 of the second 8-bit model the rates are \qty{90}{\percent} and \qty{0}{\percent}, and for channel~5 of the 4-bit model \qty{88}{\percent} and \qty{0}{\percent}. Removing this channel breaks exactly the additions that need a carry. Only 10 of the 8-bit test inputs have no carry, so their rates are coarse.

\subsection{Spatial layout effects}

Across all multi-bit experiments, placing interacting bits closer together in the spatial layout consistently reduces convergence time. In the adder experiments, the two-column layout outperforms the single-column layout, also at a fixed width (Section~\ref{sec:adders}). Our explanation, which is a hypothesis, is that corresponding operand bits are spatially adjacent in the two-column layout. This would mirror the effect of increasing the perception kernel size: both increase the amount of relevant context available to each cell in a single step, reducing the number of steps needed to propagate information across the grid.

\subsection{ALU limits and carry bit failure (single run)}

In our single run, the \ac{ALU} fails on carry output for three of eight operations. The following account is a hypothesis. ADD and SUB require a persistent carry chain, and our hypothesis is that the carry wavefront of the pure-adder model (Fig.~\ref{fig:carry_wave}) is harder to sustain under opcode conditioning, when the wavefront must stay active across bit positions while the rest of the grid routes operand data for the selected opcode. RCL does not require a chain under our definition: its carry-out is a single bit copied from the operand, so this account does not cover it, and we have no explanation for why RCL fails while its mirror RCR succeeds. Resolving carry output needs a longer training budget, a carry-preserving channel assignment, or a richer perception kernel, together with a larger evaluation sample; we leave this to future work.

\section{Conclusion}
\label{sec:conclusion}

We have shown that \acp{NCA}, under a purely local uniform update rule, can be trained to perform structured digital computation across a sequence of tasks from single gates to 8-bit ripple-carry adders, with a preliminary \ac{ALU}.

\textbf{Converged results} (10 seeds unless stated): (1) every gate and adder model reaches \qty{100}{\percent} bit correctness, and exhaustive tests find at most 7 errors in \num{65536} input pairs; (2) at a fixed width, the operand layout decides convergence; (3) adders trained on $K \geq 3$ bits extrapolate to 8 bits; (4) the carry wave has a measurable delay per chain bit and depends on a specific hidden channel.

\textbf{Limitations and hypotheses}: answers are correct within the trained rollout length but degrade beyond it. The role of $K=3$ in length generalization and the effect of operand adjacency are hypotheses. The \ac{ALU} is a single run that has not converged and learns carry-out for RCR only.

Two next steps follow: a multi-seed evaluation of the \ac{ALU}, and training for outputs that stay stable beyond the training horizon.

\section*{Acknowledgments}
This paper was prepared with the assistance of AI tools: Anthropic's Claude (several model versions)~\cite{ref_claude} and the OpenCode coding agent~\cite{ref_opencode}, used for language refinement, formatting and code assistance. The authors reviewed all AI-assisted text and code, and the experimental results, analysis, and conclusions are the authors' own work. The code and data for this study are available at \url{https://github.com/ichko/ncpu}, and an interactive demo at \url{https://ncpu.pages.dev}.

\bibliographystyle{IEEEtran}
\bibliography{references}

\end{document}